# LeViT-UNet++: Towards Balanced Fine-Grained Details and Context for Efficient Medical Image Segmentation

#Guoping Xu[1,2], Jinling Zhu[1], Liuyue Li[3]

[1]Hubei Key Laboratory of Intelligent Robot, School of Computer Science and Engineering, Wuhan Institute of Technology, Wuhan, Hubei, 430205, China

[2]Key Laboratory of Symbolic Computation and Knowledge Engineering of Ministry of Education, Jilin University, Changchun, 130012, China

[3]College of Computing and Data Science, Nanyang Technological University, Singapore, 639798, Singapore

#Corresponding Author: xugp@wit.edu.cn

**Abstract**

Efficient medical image segmentation is essential for clinical applications requiring rapid and accurate delineation, including surgical navigation, radiation therapy, and real-time image-guided intervention. Despite substantial advances in convolutional neural networks and Transformers, achieving high segmentation accuracy with limited computational resources remains challenging. In particular, reducing model complexity often comes at the expense of fine-grained structural details and effective global context modeling, resulting in an unfavorable trade-off between computational efficiency and segmentation accuracy. To address this challenge, we propose LeViT-UNet++, an efficient encoder–decoder architecture designed to achieve a better balance between local detail preservation and global context modeling. Specifically, we introduce a Haar Wavelet Multi-Scale Downsampling (HWMD) module that explicitly decomposes feature representations into low- and high-frequency components and extracts multi-scale structural information to mitigate the loss of fine-grained details during downsampling. Furthermore, we develop a depthwise convolution-based decoder that efficiently integrates multi-scale global representations from Transformer blocks with local spatial features, enabling progressive feature refinement while reducing computational overhead. We evaluate LeViT-UNet++ on four medical image segmentation benchmarks—Synapse, ACDC, LA2018, and MSD Heart—using both 2D and 3D experimental settings. Extensive experiments show that the proposed approach yields competitive segmentation accuracy with a compact parameter footprint and fast inference speed. On the Synapse dataset, for instance, LeViT-UNet++ attains a DSC of 76.57% with merely 8.025M parameters and an inference rate of 241.72 FPS, confirming its favorable accuracy–efficiency balance. This work demonstrates the effectiveness of integrating wavelet-based multi-scale feature extraction and lightweight local–global feature fusion for efficient medical segmentation. The source code will be publicly available at https://github.com/apple1986/levit_unet_pp.

## 1. Introduction

Efficient medical image segmentation is essential for clinical applications that require rapid and accurate anatomical delineation, including surgical navigation, radiation therapy, and real-time image-guided intervention [1]. Over the past decade, encoder–decoder architectures have become a dominant paradigm for medical image segmentation, with representative approaches such as U-Net [2], TransUNet [3] and Swin UNETR [4] demonstrating substantial improvements in segmentation accuracy. However, the pursuit of increasingly powerful feature representations has often been accompanied by growing computational and memory requirements, which can hinder the deployment of these models in resource-constrained and latency-sensitive clinical environments. Therefore, developing medical image segmentation networks that achieve a favorable balance between segmentation accuracy and computational efficiency remains an important research challenge, particularly for clinical applications requiring accurate and timely delineation under stringent computational constraints.

Currently, convolutional, and Transformer-based architectures have emerged as the mainstream paradigms for medical image segmentation. Convolutional architectures primarily exploit local spatial interactions to progressively extract semantic representations through hierarchical feature downsampling [5] (see Figure 1 (a)). As the spatial resolution decreases, the number of feature channels typically increases, resulting in substantial parameter and computational costs in deeper stages. Transformer-based architectures offer an alternative mechanism for modeling long-range dependencies through token-level self-attention (see Figure 1 (b)). This enables effective global contextual modeling but generally introduces substantial computational and memory costs, particularly when processing high-resolution medical images [6]. Thus, neither purely convolutional nor purely Transformer-based architectures inherently provide an optimal solution for simultaneously preserving local structural details, capturing global contextual information, and maintaining computational efficiency.

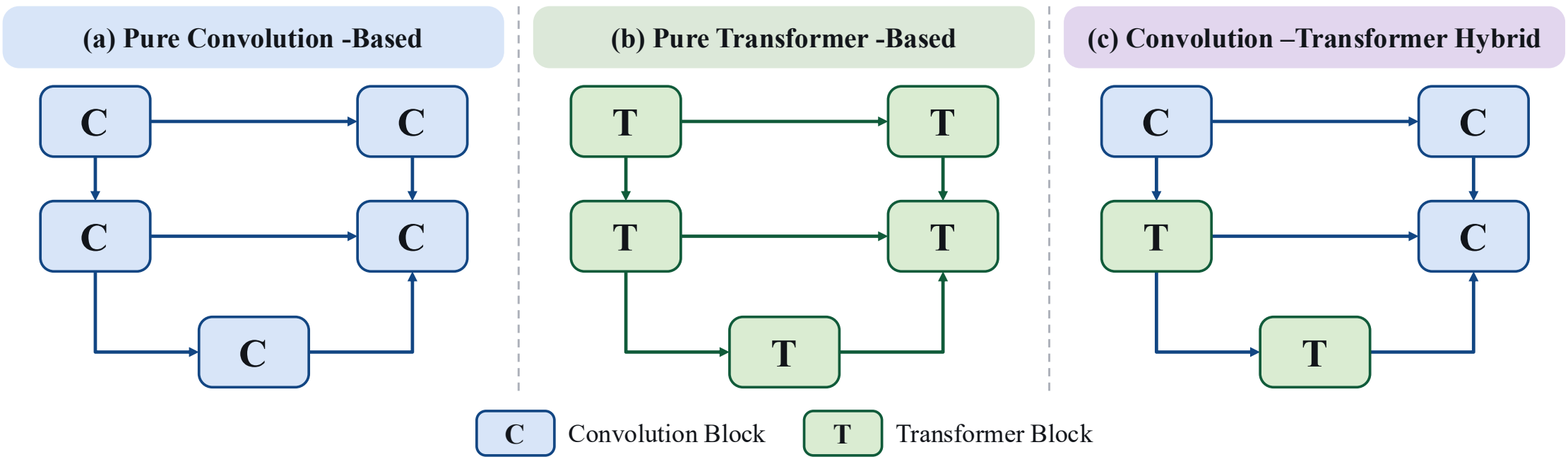

Figure 1. Three categories of segmentation architectures. (a) Pure convolution-based methods, represented by U-Net; (b) Pure Transformer-based methods, e.g., SETR [6]; (c) Convolution–Transformer hybrid methods, exemplified by LeViT-UNet [7]. Here, C denotes convolution-based blocks, and T represents Transformer-based blocks. Although the full architecture typically contains more than two such blocks, only two are shown for simplicity.

These observations motivate the exploration of hybrid architectures that exploit the complementary properties of convolution and Transformer representations. In particular, convolutional operations are well suited to capturing local spatial structures and fine-grained anatomical boundaries, whereas Transformer-based representations provide broader contextual dependencies. An effective and efficient segmentation architecture should therefore integrate these complementary representations while minimizing redundant computation during feature extraction and decoding. The LeViT-UNet [7] (see Figure 1 (c)) represents an early effort to bridge convolutional and Transformer-based representations for efficient medical image segmentation. Specifically, it adopts vanilla LeViT [8] (a lightweight hybrid convolution–Transformer backbone originally designed for image classification) as the encoder and convolution-based blocks as the decoder, seeking to achieve a favorable trade-off between segmentation accuracy and computational efficiency. Its core design is based on the observation that convolutional operations can efficiently extract local spatial representations, while progressive downsampling reduces the spatial resolution of feature maps. Transformer blocks can subsequently model long-range dependencies on these low-resolution representations with substantially lower computational cost than applying global attention at high spatial resolutions. This hybrid design therefore provides an efficient means of combining local feature extraction with global contextual modeling for medical image segmentation.

However, LeViT-UNet-like architectures [3, 7, 9] still face two important limitations. First, their efficiency-oriented convolutional encoders rely heavily on progressive spatial downsampling to reduce computational cost and expand the effective receptive field. Although this strategy enables efficient hierarchical feature extraction, conventional downsampling operations, such as max pooling and average pooling, inevitably discard spatial information [10]. This information loss is particularly problematic for medical image segmentation, where accurate delineation often depends on subtle boundary cues, thin structures, and small anatomical regions. In particular, high-frequency components contain important fine-grained structural information that may be attenuated or lost during repeated spatial reduction [11]. Once these details are discarded at early stages, subsequent upsampling operations cannot fully reconstruct the original structural information. Therefore, an efficient segmentation architecture should not rely solely on spatial downsampling for computational reduction, but should also provide an explicit mechanism for preserving complementary fine-grained structural information throughout the feature extraction process.

Second, convolution-based decoders suffer from notable computational redundancy in feature learning [12, 13]. Within the standard encoder–decoder architecture, the decoder gradually fuses features from its previous stage with skip-connected features from the matching encoder layer. Although skip connections help restore spatial details, their fusion commonly employs standard convolutions over high-dimensional feature maps. This adds a large number of convolutional parameters and computation overhead in the decoding branch. Furthermore, concatenated encoder and decoder features often contain overlapping, redundant spatial representations, where not all signals are essential for faithful feature reconstruction. This motivates a core research question: *can the decoder maintain the representational capacity of conventional* *convolutional feature fusion while drastically cutting its computational redundancy?*

To address these two challenges, we introduce two complementary architectural enhancements to preserve fine-grained structural information during encoder downsampling and improve computational efficiency in the decoder. First, we propose a Haar Wavelet Multi-Scale Downsampling (HWMD) module that combines Haar wavelet decomposition with parallel depthwise convolutions of different kernel sizes. Haar wavelet decomposition separates feature representations into low- and high-frequency components, capturing complementary structural information, while multi-scale depthwise convolutions extract spatial patterns across different receptive fields with limited additional computational cost. This design helps preserve fine-grained anatomical details during downsampling while maintaining efficient feature extraction.

Furthermore, we develop a Depthwise-based Local and Global feature fusion and learning decoder (DLGD) that employs lightweight depthwise convolutional blocks [14] to progressively integrate global contextual representations from Transformer blocks with local spatial features propagated through encoder skip connections. By enabling efficient feature fusion and refinement, the proposed decoder reduces computational overhead while integrating complementary local and global representations. Together, HWMD and DLGD improve the balance between structural detail preservation, contextual representation, and computational efficiency.

In summary, we introduce LeViT-UNet++, an enhanced version of our previous LeViT-UNet [7] that improves both the encoding and decoding processes for efficient medical image segmentation. The main contributions of this study are as follows:

(1) We introduce a Haar Wavelet Multi-scale Downsampling module (HWMD) that combines wavelet-based information preservation with parallel depthwise convolutions at multiple scales. HWMD retains complementary structural information during spatial downsampling while providing multi-scale representations with limited computational overhead for the encoder.

(2) We propose a Depthwise-based Local and Global feature fusion and learning decoder (DLGD) that efficiently integrates local spatial features from the convolutional encoder with global contextual representations from the Transformer. By leveraging lightweight depthwise convolutional blocks for progressive feature fusion and refinement, DLGD reduces decoder computational overhead while effectively integrating complementary local and global information.

(3) We develop LeViT-UNet++, a convolution–Transformer hybrid encoder–decoder framework that integrates HWMD and DLGD to jointly preserve fine-grained structural information, capture global contextual representations, and improve computational efficiency.

(4) We conduct extensive experiments on four challenging medical image segmentation benchmarks covering both 2D and 3D segmentation scenarios. The results demonstrate that LeViT-UNet++ achieves a competitive balance between segmentation performance and computational efficiency compared with representative state-of-the-art methods.

## 2. Related Work

### 2.1 Downsampling in Image Segmentation

Downsampling is a fundamental operation in semantic segmentation networks, as it progressively reduces spatial resolution to decrease computational cost and enlarge the effective receptive field while constructing hierarchical feature representations. Conventional segmentation architectures typically employ pooling operations, such as max pooling [15] and average pooling [16], for spatial reduction. For example, SegNet [15] adopts pooling indices to facilitate efficient feature reconstruction in the decoder. Beyond conventional pooling, multi-scale feature aggregation has been widely explored to improve the representation of objects with varying spatial scales. PSPNet [17] introduces spatial pyramid pooling to aggregate contextual information from multiple spatial regions, while DeepLab [18] employs atrous spatial pyramid pooling to capture multi-scale contextual representations without further reducing spatial resolution.

Despite their effectiveness, conventional downsampling operations inevitably discard spatial information as the feature resolution decreases. Such information loss can affect segmentation performance, particularly for objects with complex boundaries, small structures, or fine-grained details. To alleviate this issue, several studies have explored information-preserving downsampling strategies [10, 11, 19]. Haar wavelet-based downsampling (HWD) [11], for instance, replaces conventional spatial reduction with Haar wavelet decomposition to preserve complementary frequency components during downsampling. By retaining low- and high-frequency information, wavelet-based approaches provide richer representations for subsequent feature extraction while maintaining the computational benefits of spatial reduction.

However, HWD employs standard convolution to process and fuse the decomposed frequency components, introducing additional learnable parameters and computational overhead compared with conventional pooling operations. Moreover, its feature extraction remains based on a fixed convolutional receptive field and does not explicitly exploit spatial patterns at multiple scales, which may limit its ability to capture structural variations across different object sizes and spatial contexts. These limitations motivate the development of a more efficient information-preserving downsampling strategy that can simultaneously retain fine-grained structural information and capture multi-scale spatial patterns. Hence, we introduce an efficient Haar Wavelet Multi-Scale Downsampling module for image segmentation to address these challenges in this study.

### 2.2 Architectures for Efficient Neural Networks

A broad range of strategies has been explored to develop lightweight neural architectures, with existing approaches generally falling into two complementary categories: efficient building-block design and holistic architectural optimization [20]. The former aims to reduce computational complexity at the individual module level, whereas the latter seeks to improve efficiency through the design and optimization of the overall network topology.

At the module level, depthwise separable convolution is a fundamental technique for reducing computational cost. It decomposes a standard convolution into a depthwise convolution, which independently processes each input channel, followed by a pointwise 1×1 convolution, which facilitates cross-channel information integration. This factorization substantially reduces the number of parameters and floating-point operations (FLOPs), particularly in networks with high-dimensional feature representations. Representative architectures that exploit this principle include the MobileNet series [13, 21, 22] and Xception [14]. In addition, ShuffleNet [23] and ShuffleNet V2 [24] combine group convolution with channel shuffling to reduce computational redundancy while enabling information exchange across channel groups. GhostNet [12] further improves computational efficiency through its Ghost module, which generates additional feature maps from a limited set of intrinsic features using inexpensive linear transformations, thereby reducing reliance on computationally intensive conventional convolutions. Collectively, these approaches demonstrate how carefully designed computational primitives can substantially improve the efficiency of deep neural networks without fundamentally compromising their representational capacity.

In contrast, holistic architectural optimization targets the overall network structure, including the allocation of computational resources across stages, the scaling of network depth, width, and input resolution, and the organization of feature extraction pathways. EfficientNet [25, 26] systematically investigates the joint scaling of these dimensions through compound model scaling, achieving a more balanced trade-off between computational cost and representational capacity. Other representative architectures, such as BiSeNet [16], DDRNet [27], and LGNet [28], explore complementary two-branch strategies for learning and fusing local and contextual information. SegFormer [29], by contrast, designs efficient MLP-based decoders to aggregate multi-layer features extracted from the encoder. Instead of optimizing individual operations in isolation, these approaches emphasize coordinated design of network components and their global configurations to attain more favorable efficiency–accuracy trade-offs.

While these lightweight architectural design principles were originally developed for natural image recognition, they have been increasingly adapted to medical image segmentation. Typical approaches include embedding efficient modules into existing networks or directly adopting lightweight backbones. Nevertheless, medical segmentation poses unique challenges: it demands a delicate balance among computational efficiency, multi-scale contextual modeling, and the preservation of fine-grained anatomical details. Representative architectures, including LeViT-UNet [7], MedNeXt [30], Mobile U-ViT [31], explore different strategies for efficient feature extraction and contextual modeling. Despite these advances, the design of an effective and computationally efficient decoder remains a critical challenge. A well-designed decoder must progressively recover spatial details and integrate complementary local and global representations across multiple scales, while minimizing the computational and memory overhead associated with feature fusion and refinement. This challenge becomes even more pronounced in 3D medical image segmentation, where the high dimensionality of volumetric features substantially increases computational demands [32].

Therefore, developing lightweight decoders that effectively balance multi-scale feature fusion, anatomical detail preservation, and computational efficiency is essential for both 2D and 3D medical image segmentation.

## 3. Method

### 3.1 Overall framework for the proposed LeViT-UNet++

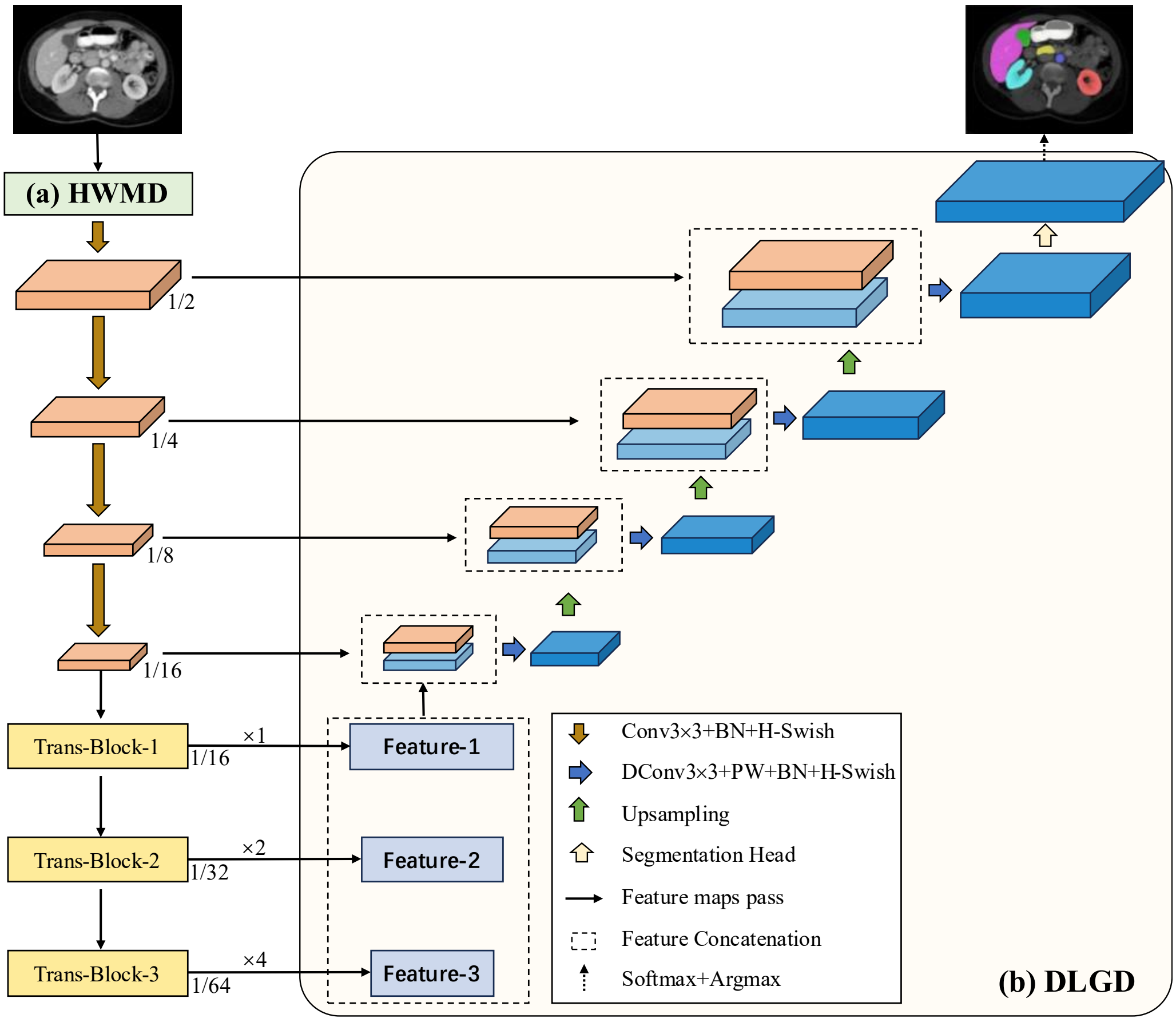

Figure 2. Overview of the proposed LeViT-UNet++, an encoder–decoder architecture that combines convolutional operations and Transformer modules. Specifically, the Haar Wavelet Multi-Scale Downsampling (HWMD) module serves as the stem of the encoder, while the Depthwise-based Local-and-Global feature fusion and learning decoder (DLGD) leverages depthwise-dominant convolutions to enable efficient feature representation learning.

The overall architecture of the proposed LeViT-UNet++ is illustrated in Figure 2. The network adopts a hybrid encoder–decoder architecture that combines convolutional operations with Transformer-based global context modeling to capture complementary local spatial details and long-range contextual dependencies. Compared with its predecessor, LeViT-UNet, the proposed architecture introduces two key enhancements. (1) A Haar Wavelet Multi-Scale Downsampling (HWMD) module is incorporated as the encoder stem to extract multi-scale structural information from low- and high-frequency components, helping preserve fine-grained details during spatial downsampling. (2) A Depthwise-based Local and Global feature fusion and learning decoder (DLGD) is designed to progressively integrate and refine local spatial features from the convolutional encoder and global contextual representations from Transformer blocks, thereby reducing decoder computational overhead. The detailed components of the encoder and decoder are listed in **Table 1**.

**Table 1.** Overall architecture of 2D LeViT-UNet++. Here, #InCh denotes the number of input feature channels, and #OutCh refers to the corresponding number of output feature channels. Input and Output indicate the spatial resolution of feature maps. Cls represents the number of target segmentation classes. ConvB, TransB, MSFusion and DConvB denote the standard convolution block, Transformer block, multi-scale feature fusion block and depthwise convolution block, respectively.

| Type | Operator | Params (M) | #InCh | #OutCh | Input | Output |
|---|---|---|---|---|---|---|
| Encoder | HWMD | 0.0002 | 1 | 16 | 224×224 | 112×112 |
| | ConvB-1 | 0.0023 | 16 | 16 | 112×112 | 112×112 |
| | ConvB-2 | 0.0047 | 16 | 32 | 112×112 | 56×56 |
| | ConvB-3 | 0.0186 | 32 | 64 | 56×56 | 28×28 |
| | ConvB-4 | 0.0740 | 64 | 128 | 28×28 | 14×14 |
| | TransB-1 | 0.2340 | 128 | 128 | 14×14 | 14×14 |
| | TransB-2 | 1.7348 | 128 | 256 | 14×14 | 7×7 |
| | TransB-3 | 4.9390 | 256 | 384 | 7×7 | 4×4 |
| Decoder | MSFusion | - | 768 | 768 | 14×14, 7×7, 4×4 | 14×14 |
| | DConvB-1 | 0.7356 | 896 | 512 | 14×14 | 28×28 |
| | DConvB-2 | 0.2215 | 576 | 256 | 28×28 | 56×56 |
| | DConvB-3 | 0.0575 | 288 | 128 | 56×56 | 112×112 |
| | SegHead | 0.0052 | 128 | Cls | 112×112 | 224×224 |

Specifically, the *encoder* consists of three sequential components: (1) a Haar Wavelet Multi-Scale Downsampling (HWMD) module, (2) four convolutional blocks, and (3) three Transformer blocks. First, HWMD reduces the spatial resolution of the input feature maps while aggregating multiscale structural information through Haar wavelet decomposition and parallel multi-scale spatial feature extraction. The resulting features are subsequently processed by four convolutional blocks, each incorporating convolutional layers, batch normalization, and hard-swish activation to extract local spatial representations. To progressively reduce computational complexity while expanding the effective receptive field, spatial downsampling is performed between consecutive blocks, except before the first convolutional block. Finally, the three successive Transformer blocks operate on the progressively downsampled feature representations to model long-range dependencies among spatial tokens, thereby enhancing global contextual modeling and complementing the local features extracted by the convolutional stages.

Subsequently, the *decoder* is elaborately designed to progressively recover spatial resolution and fuse complementary multi-level feature hierarchies delivered by the encoder. It consists of a multi-scale global feature fusion module, three sequentially arranged depthwise separable convolutional blocks within the upsampling pathway, and a final segmentation head. The fusion module aggregates hierarchical feature representations derived from the three Transformer blocks to achieve effective cross-scale information integration and strengthen global contextual consistency. The integrated global features are further concatenated with local feature maps generated by the last convolutional block of the encoder. The concatenated global-local features are then progressively refined via three successive depthwise separable convolutional layers, each followed by a 2× upsampling operation. This design gradually restores fine spatial details while substantially reducing computational burden and parameter overhead. Compared with vanilla LeViT-UNet, our proposed DLGD decoder reduces the parameter count by nearly 49.5%. Ultimately, the segmentation head adopts a convolutional layer combined with bilinear upsampling to produce high-resolution pixel-wise segmentation predictions.

### 3.2 Haar Wavelet Multi-Scale Downsampling Module

The proposed Haar Wavelet Multi-Scale Downsampling (HWMD) module is designed to preserve fine-grained structural information while progressively reducing spatial resolution. HWMD explicitly decomposes the input feature map into low- and high-frequency components using the two-dimensional discrete Haar wavelet transform, extracts complementary representations through multi-scale depthwise convolutions, and subsequently integrates these representations using pointwise convolution. As illustrated in Figure 2 (a), HWMD consists of three successive operations: Haar wavelet decomposition, frequency-specific multi-scale depthwise convolution, and low–high-frequency feature fusion.

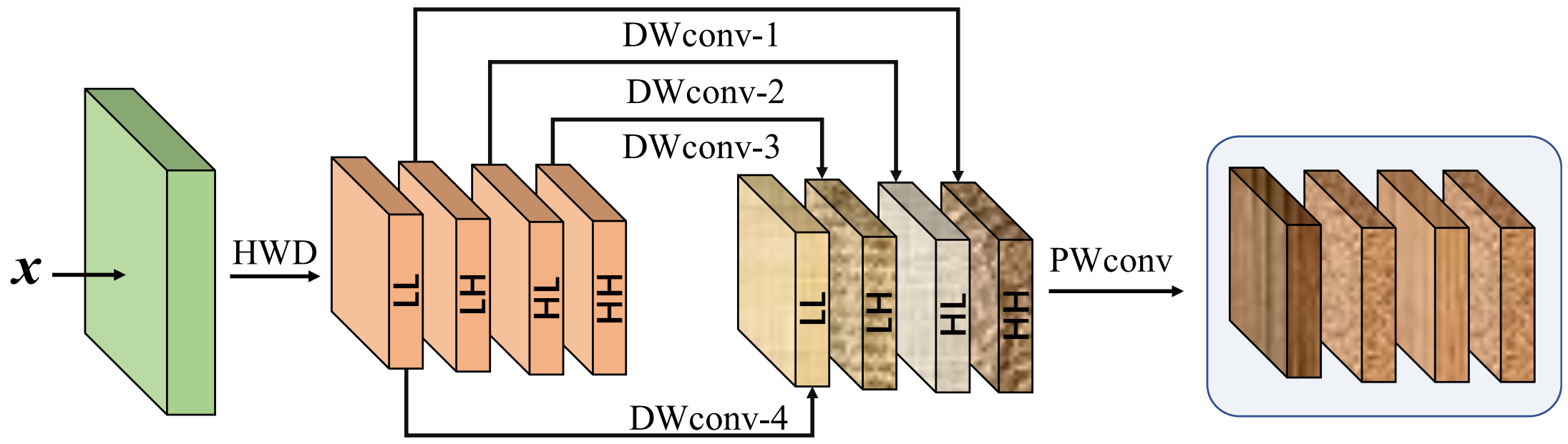

Figure 3. Illustration of the proposed HWMD module, which consists of Haar wavelet decomposition (HWD), frequency-aware multi-scale depthwise convolution, and low–high-frequency feature fusion with pointwise convolution.

(1) Haar wavelet transformation

Given an input feature map X with dimensions H × W × C, where H, W, and C denote the spatial height, spatial width, and number of channels, respectively, the two-dimensional Haar wavelet transform decomposes each channel into four frequency subbands. Assuming even spatial dimensions, the transform can be expressed as follows.

$$
\begin{aligned}
X_{LL} &= \frac{1}{2}(X_{00} + X_{01} + X_{10} + X_{11}) \\
X_{LH} &= \frac{1}{2}(X_{00} - X_{01} + X_{10} - X_{11}) \\
X_{HL} &= \frac{1}{2}(X_{00} + X_{01} - X_{10} - X_{11}) \\
X_{HH} &= \frac{1}{2}(X_{00} - X_{01} - X_{10} + X_{11})
\end{aligned} \tag{1}
$$

Here, $X_{00}$, $X_{01}$, $X_{10}$, and $X_{11}$represent the four spatial samples within each non-overlapping 2 × 2 neighborhood of the input feature map. The LL subband captures low-frequency information associated primarily with coarse structures and global appearance, whereas the LH, HL, and HH subbands encode high-frequency variations along different spatial orientations. The resulting subbands have spatial dimensions H/2 × W/2, thereby achieving a twofold reduction along each spatial dimension while explicitly retaining complementary frequency information.

(2) Frequency-specific multi-scale depthwise convolution

To extract features at different receptive-field scales while maintaining computational efficiency, HWMD applies parallel depthwise convolutions with different kernel sizes to the low- and high-frequency components, respectively. Let K = ($k_1$, $k_2$, ..., $k_m$) denote the set of convolutional kernel sizes. The low-frequency representation is obtained as follows.

$$F_L = Concat_{(k \in K)} DWconv_{(k \times k)}(X_{LL}) \tag{2}$$

Here, the *Concat* denotes concatenation along the channel dimension. This operation captures low-frequency contextual patterns over multiple receptive fields. For the high-frequency branch, the three directional subbands are first aggregated along the channel dimension.

$$X_H = Concat(X_{LH}, X_{HL}, X_{HH}) \tag{3}$$

The aggregated high-frequency representation is then processed using the corresponding multi-scale depthwise convolutions.

$$F_H = Concat_{(k \in K)} DWConv_{(k \times k)}(X_H) \tag{4}$$

By operating independently on the low- and high-frequency representations, the two branches learn complementary features: the low-frequency branch emphasizes coarse semantic structures, whereas the high-frequency branch captures fine-grained variations and orientation-dependent structural cues. The parallel multi-scale design further enables the module to characterize structures of different spatial extents while limiting the computational overhead associated with conventional dense convolutions.

(3) Low–high-frequency feature fusion

Finally, the extracted low- and high-frequency representations are integrated using a pointwise convolution. Specifically, the two branches are concatenated and projected into a unified feature representation.

$$Y = PWconv_{(1\times1)}\big(Concat(F_L, F_H)\big) \tag{5}$$

Here, Y denotes the output feature map with dimensions H/2 × W/2 × C_out, where C_out is the desired output channel dimension. The 1 × 1 convolution enables channel-wise interactions between complementary frequency representations and integrates multi-scale contextual information with fine-grained structural details. Consequently, HWMD performs spatial downsampling and frequency-aware feature extraction within a unified module, providing a computationally efficient alternative to conventional downsampling operations that may inadequately preserve fine structural information.

### 3.3 Depthwise-based local-and-global feature fusion and learning decoder

The proposed decoder (DLGD) is designed to efficiently integrate multi-scale global contextual representations from the Transformer encoder with local spatial features, progressively refining the fused representations to recover fine-grained details for accurate segmentation. As illustrated in Figure 2 (b), the decoder consists of three principal components: (1) a multi-scale global feature fusion module that aggregates hierarchical representations from the Transformer blocks; (2) a local and global feature fusion and learning module comprising three depthwise separable convolutional blocks, which progressively fuse and refine complementary local and global features while maintaining computational efficiency; and (3) a segmentation head that generates the final pixel-wise prediction.

Specifically, the multi-scale global feature fusion module combines hierarchical Transformer features to provide contextual guidance at different semantic levels. These fused global representations are subsequently integrated with local spatial features through the three successive depthwise separable convolutional blocks. By progressively refining the fused features, DLGD facilitates the interaction between global semantic context and local spatial information, thereby supporting the recovery of fine-grained anatomical structures and object boundaries. Furthermore, the use of depthwise separable convolutions reduces the computational cost of feature refinement compared with conventional convolutional operations, making the decoder more parameter-efficient. Overall, DLGD provides a lightweight decoding strategy that balances multi-scale contextual integration, local detail recovery, and computational efficiency.

## 4. Experiments and Results

### 4.1 Datasets

(1). Synapse dataset

The Synapse Multi-Organ Segmentation (Synapse) dataset contains 30 abdominal CT volumes, offering dense voxel-level annotations for eight abdominal organs: aorta (Aor), gallbladder (Gal), spleen (Spl), left kidney (LKid), right kidney (RKid), liver (Liv), pancreas (Pan), and stomach (Sto). We adopt the data split strategy described in [7], where 18 samples are allocated for model training and the other 12 samples are held out as the validation set.

(2) ACDC

The Automated Cardiac Diagnosis Challenge (ACDC) dataset collects cardiac cine-MRI sequences captured at end-diastole and end-systole. Every subject is annotated manually for three key cardiac structures, namely the left ventricle (LV), myocardium (MYO), and right ventricle (RV), to construct pixel-wise ground truth. Consistent with the splitting scheme reported in [7], we select 20 subjects (40 volumetric scans) as the test set, and utilize the remaining 80 subjects (160 volumes) for network training.

(3) LA2018

The Left Atrium Segmentation Challenge (LA2018) dataset [33] comprises 100 volumetric late gadolinium-enhanced magnetic resonance imaging (LGE-MRI) scans, each accompanied by a manually delineated segmentation mask of the left atrium (LA). In this study, the dataset is randomly partitioned into 70 volumes for training, 10 for validation, and 20 for testing.

(4) MSD Heart

The Medical Segmentation Decathlon (MSD) Heart dataset (Task02_Heart) [34] provides volumetric cardiac magnetic resonance imaging (MRI) data with segmentation annotations for the target cardiac structure. The dataset contains 30 cases, including 20 cases with ground-truth segmentation labels and 10 cases without publicly available labels. In this study, the 20 labeled cases are randomly partitioned into 14 training cases, 2 validation cases, and 4 testing cases to evaluate the performance of the proposed method and competing three-dimensional (3D) segmentation models.

**Table 2** summarizes the four publicly available benchmark datasets used in this study, detailing their imaging modalities, target anatomical structures, and respective training, validation, and test splits. Different data preparation and evaluation protocols are adopted for the 2D- and 3D-based segmentation models. Specifically, for the Synapse and ACDC datasets, volumetric scans are decomposed into individual 2D slices for model training and inference. In contrast, the original volumetric data from the LA2018 and MSD Heart datasets are retained for training and evaluating 3D segmentation models. To ensure a consistent volume-level evaluation across different model architectures, the slice-wise predictions generated by the 2D models are subsequently reassembled into 3D volumes in their original spatial order. All segmentation metrics are then computed at the volumetric level, providing a unified basis for comparing segmentation performance across the evaluated models.

**Table 2.** Summary of the four public datasets used for training and evaluation.

| Dataset | Modality | Objects | Train | Val | Test |
|---|---|---|---|---|---|
| Synapse | CT | Abdominal Organ | 18 | 0 | 12 |
| ACDC | MR | Cardiac | 160 | 0 | 40 |
| LA2018 | MR | Left Atrium | 70 | 10 | 20 |
| MSD Heart | MR | Heart | 14 | 2 | 4 |

For the Synapse and ACDC datasets, segmentation performance is assessed using the Dice similarity coefficient (DSC) and the 95th-percentile Hausdorff distance (HD95), following the standard evaluation protocols adopted in previous studies [3, 7]. For the LA2018 and MSD Heart datasets, two additional surface-based metrics, namely the normalized surface dice (NSD) and average surface distance (ASD), are employed to provide a more comprehensive assessment of boundary delineation accuracy. Specifically, NSD quantifies the proportion of predicted and reference surface points falling within a predefined tolerance, thereby reflecting boundary agreement, whereas ASD measures the average distance between the predicted and reference segmentation surfaces, characterizing the overall boundary displacement.

### 4.2 Implementation details

In this study, all experiments are implemented using Python 3.10.4 and PyTorch 2.11 with CUDA 12.8 on an NVIDIA GeForce RTX 4090 GPU, running under Ubuntu 24.04.4 LTS. For all training samples, augmentation strategies including random flipping and random rotation are adopted to enhance data diversity. For 2D model training, all input images are resized to a spatial resolution of 224 × 224. All 2D models are optimized with the Adam optimizer, using an initial learning rate of 5e-4 and weight decay of 1e-4. Cross-entropy loss combined with Dice loss serves as the objective function. The batch size is set to 8. We train the model for 350 epochs on the Synapse dataset and 400 epochs on the ACDC dataset. Following the strategy adopted in [3, 7], we select the final epoch for evaluation on the test set.

For 3D model training, we use the MONAI framework and adopt the experimental configurations of Swin UNETR [4] as a reference for the LA2018 and MSD Heart datasets. Specifically, we extend the 2D HWMD and DLGD modules to their 3D counterparts based on 3D Haar wavelet transforms and convolutions for the 3D LeViT-UNet++. AdamW is used as the optimizer, with an initial learning rate of 0.001 and a weight decay of $1 \times 10^{-5}$. The training objective combines cross-entropy loss and Dice loss. The input volume size is obtained by randomly cropping 64 × 64 × 64 voxel patches centered on target regions, and the batch size is set to 4. The models are trained for 200 epochs on LA2018 and 600 epochs on MSD Heart, respectively. During training, model selection is performed based on the mean Dice score on the validation set, and the checkpoint achieving the highest validation Dice score is retained for final evaluation on the corresponding test set. For methods with multiple publicly available variants, the smallest model in terms of parameter count is selected to ensure a consistent comparison of model efficiency.

### 4.3 Experiment results under 2D setting

(1) Results on Synapse dataset

We compare our proposed LeViT-UNet++ against a collection of state-of-the-art (SOTA) baselines with competitive segmentation accuracy and computational efficiency. The compared methods encompass convolution-based models including CMUNeXt [35], BiSeNetV2 [16] and DDRNet [27]; transformer-dominant architectures such as SegFormer [29] and Swin-UNet [36]; as well as hybrid architectures comprising EfficientViT [37],

MobileU-ViT [31], UNeXt [38], Rolling-UNet [39], LeViT-UNet [7], H2Former [40], TransUNet [3]and UCTransNet [41]. Quantitative results in terms of the Dice similarity coefficient (DSC) and 95% Hausdorff Distance (HD95) are reported in **Table 3** (a) and (b).

**Table 3.** Quantitative comparison of segmentation performance and computational efficiency on the Synapse multi-organ dataset. We report the model type, number of parameters (Params, in M), inference speed (FPS), the averaged DSC and HD95, as well as per-organ DSC values for Aorta (Aor), Gallbladder (Gal), Left Kidney (Lkid), Right Kidney (Rkid), Liver (Liv), Pancreas (Pan), Spleen (Spl), and Stomach (Sto). ↓ indicates lower is better; ↑ indicates higher is better. The same notation applies to all subsequent tables. The abbreviations C, T, C+T, and C+MLP denote convolution-based, Transformer-based, convolution–Transformer hybrid, and convolution–multilayer perceptron hybrid architectures, respectively.

**(a) Model efficiency and overall performance measured by DSC (%) and HD95.**

| Networks | Type | Params (M) ↓ | FPS ↑ | DSC ↑ | HD95 ↓ |
|---|---|---|---|---|---|
| LeViT-UNet++ | C+T | 8.025 | 241.72 | 76.57 | 18.29 |
| CMUNeXt | C | 0.417 | 1029.18 | 74.46 | 38.12 |
| EfficientViT | C+T | 0.704 | 993.70 | 70.38 | 35.14 |
| Mobile U-ViT | C+T | 1.390 | 243.79 | 79.46 | 28.48 |
| UNeXt | C+MLP | 1.472 | 1023.60 | 71.24 | 33.79 |
| Rolling-Unet | C+MLP | 1.783 | 144.24 | 79.98 | 25.37 |
| BiSeNetV2 | C | 3.618 | 739.11 | 75.50 | 20.01 |
| SegFormer | T | 3.712 | 240.91 | 73.31 | 25.44 |
| DDRNet | C | 5.694 | 692.33 | 76.03 | 19.84 |
| Swin-UnetR | C+T | 6.302 | 173.90 | 75.22 | 28.89 |
| LeViT-UNet | C+T | 15.896 | 237.31 | 73.69 | 23.92 |
| Swin-Unet | T | 27.165 | 142.64 | 62.63 | 49.17 |
| H2Former | C+T | 33.667 | 160.32 | 80.54 | 14.55 |
| TransUNet | C+T | 67.005 | 204.45 | 75.31 | 42.87 |
| UCTransNet | C+T | 66.430 | 42.20 | 80.09 | 25.09 |

**(b) Per-organ segmentation performance in terms of DSC (%)**

| Networks | Aor | Gal | Lkid | Rkid | Liv | Pan | Spl | Sto |
|---|---|---|---|---|---|---|---|---|
| LeViT-UNet++ | 85.28 | 59.17 | 84.07 | 79.33 | 93.60 | 55.47 | 86.76 | 68.91 |
| CMUNeXt | 86.62 | 63.12 | 81.23 | 69.38 | 93.11 | 46.21 | 87.31 | 68.69 |
| EfficientViT | 80.81 | 55.29 | 69.29 | 63.47 | 92.20 | 47.86 | 84.87 | 69.22 |
| Mobile U-ViT | 87.19 | 61.45 | 84.60 | 81.87 | 93.39 | 59.05 | 90.52 | 77.65 |
| UNeXt | 82.33 | 56.52 | 71.63 | 68.54 | 91.98 | 46.11 | 84.23 | 68.61 |
| Rolling-Unet | 87.49 | 66.34 | 84.25 | 80.06 | 94.69 | 59.75 | 90.65 | 76.63 |
| BiSeNetV2 | 83.42 | 65.50 | 82.45 | 73.45 | 92.58 | 53.14 | 82.15 | 71.30 |
| SegFormer | 81.26 | 56.99 | 80.27 | 76.19 | 92.30 | 49.21 | 82.80 | 67.48 |
| DDRNet | 83.22 | 62.09 | 79.81 | 77.45 | 92.53 | 51.30 | 89.59 | 72.27 |
| Swin-UnetR | 86.60 | 61.88 | 81.61 | 70.66 | 93.06 | 53.89 | 86.50 | 67.55 |

| Networks | Aor | Gal | Lkid | Rkid | Liv | Pan | Spl | Sto |
|---|---|---|---|---|---|---|---|---|
| LeViT-UNet | 86.45 | 66.13 | 79.32 | 73.56 | 91.85 | 49.25 | 79.29 | 63.70 |
| Swin-Unet | 63.91 | 55.66 | 61.87 | 62.75 | 88.85 | 31.61 | 78.04 | 58.37 |
| H2Former | 88.44 | 68.99 | 86.17 | 83.96 | 94.30 | 60.39 | 91.72 | 70.34 |
| TransUNet | 88.09 | 59.09 | 82.06 | 75.09 | 91.62 | 54.59 | 85.13 | 66.78 |
| UCTransNet | 88.60 | 70.45 | 83.94 | 78.88 | 94.06 | 64.06 | 87.54 | 73.20 |

As shown in **Table 3**, our LeViT-UNet++ achieves a competitive overall DSC of 76.57% and HD95 of 18.29, with only 8.03M parameters and an inference FPS of 242. Compared with its predecessor LeViT-UNet, our LeViT-UNet++ obtains a clear performance gain (+2.88% overall DSC) while reducing the parameter count from 15.90M to 8.03M and maintaining comparable inference speed. Among all models with less than 10M parameters, LeViT-UNet++ achieves the best HD95 metric and competitive overall DSC. Specifically, our method yields strong results on the Aorta, Liver and Spleen. Although heavy-weight hybrid models such as H2Former and UCTransNet deliver higher overall DSC, they suffer from substantially larger parameter overhead. Meanwhile, ultra-light CNN baselines like CMUNeXt run much faster but exhibit obvious degradation on HD95, showing inferior boundary segmentation capability.

(2) Results on ACDC dataset

We further validate the performance of our LeViT-UNet++ on the ACDC cardiac segmentation dataset. As shown in **Table 4**, we compare the proposed model with the same set of lightweight SOTA baselines adopted in the Synapse experiment, covering pure CNN, transformer-only and hybrid CNN-transformer/CNN-MLP architectures.

**Table 4.** Quantitative segmentation results on the ACDC cardiac dataset. We report the mean Dice similarity coefficient (Dice), 95% Hausdorff Distance (HD95), and per-structure Dice for the right ventricle (RV), myocardium (MYO), and left ventricle (LV).

| Networks | DSC↑ | HD95↓ | RV | MYO | LV |
|---|---|---|---|---|---|
| LeViT-UNet++ | 88.95 | 1.35 | 88.05 | 86.16 | 92.62 |
| CMUNeXt | 90.93 | 1.16 | 90.07 | 88.57 | 94.14 |
| EfficientViT | 87.70 | 2.13 | 87.24 | 84.27 | 91.59 |
| Mobile U-ViT | 91.26 | 1.15 | 90.90 | 88.67 | 94.23 |
| UNeXt | 89.57 | 1.45 | 89.02 | 86.64 | 93.06 |
| Rolling-Unet | 91.64 | 1.11 | 90.96 | 89.23 | 94.73 |
| BiSeNetV2 | 90.40 | 1.48 | 90.62 | 87.49 | 93.10 |
| SegFormer | 87.70 | 2.13 | 87.24 | 84.27 | 91.59 |
| DDRNet | 90.35 | 1.60 | 90.04 | 87.67 | 93.33 |
| Swin-UnetR | 90.63 | 1.50 | 90.49 | 88.13 | 93.27 |
| LeViT-UNet | 89.39 | 1.30 | 88.16 | 86.97 | 93.05 |
| Swin-Unet | 81.84 | 3.86 | 80.85 | 78.87 | 85.81 |
| H2Former | 91.33 | 1.11 | 90.54 | 88.79 | 94.67 |
| TransUNet | 91.14 | 1.12 | 90.47 | 88.59 | 94.35 |
| UCTransNet | 91.71 | 1.06 | 91.31 | 89.13 | 94.69 |

Our LeViT-UNet++ achieves an overall Dice of 88.95% and HD95 of 1.35. Compared

with its predecessor LeViT-UNet, our model obtains competitive overall segmentation accuracy and comparable HD95. For the three cardiac substructures, LeViT-UNet++ achieves Dice values of 88.05% for RV, 86.16% for MYO and 92.62% for LV. The top-performing methods such as UCTransNet and H2Former achieve slightly higher Dice and better HD95, but these models come with much heavier computational overhead as shown in **Table 3** (a). On the other hand, many lightweight baselines like EfficientViT and SegFormer exhibit obvious performance drops, especially for HD95. LeViT-UNet++ maintains balanced segmentation quality on cardiac structures with a lightweight parameter budget, demonstrating strong generalization capability across abdominal multi-organ and cardiac segmentation tasks.

### 4.4 Experiment results under 3D setting

(1) Results on LA2018 dataset

To evaluate the performance of LeViT-UNet++, we conduct experiments under the 3D volume setting by comparing it with a variety of state-of-the-art 3D segmentation methods: convolution-based models such as MedNeXt [30], UNet++, nnUNet, V-Net [42] and UXNet [43], Transformer-based methods including SegFormer3D and nnFormer [44], and hybrid architectures represented by Mobile U-ViT, LeViT-UNet and Swin UNETR.

As shown in **Table 5**, the proposed LeViT-UNet++ achieves a competitive DSC of 84.88%, while maintaining a small parameter footprint of only 8.27 M and a favorable inference FPS of 203.52. it reaches the second-lowest ASD (6.32) and a competitive NSD (23.22), demonstrating reliable boundary segmentation. MedNeXt-v1 and V-Net obtain marginally higher DSC (85.05% and 85.82%), but V-Net has 45.60 M parameters, over five times our model’s size. nnUNet runs fastest at 231.99 FPS, but its DSC (82.46%) and boundary metrics are inferior. nnFormer has the lowest NSD (14.96), indicating weak boundary agreement, and it is computationally expensive (149.46 M params, 21.81 FPS). Relative to LeViT-UNet, our method boosts DSC from 81.50% to 84.88% and reduces parameters by 75.6%, with improved HD95 and NSD. SegFormer3D achieves a similar DSC (84.16%) but poorer ASD and HD95; Mobile U-ViT is lightweight (2.76 M) yet underperforms on DSC and boundary metrics. In summary, LeViT-UNet++ strikes a good trade-off between segmentation accuracy, boundary quality, parameter cost and inference efficiency.

**Table 5.** Quantitative comparison of segmentation performance and computational efficiency on the LA2018 dataset. We report the model type, number of parameters (Params, in M), inference speed (FPS), as well as the averaged DSC (%), HD95 (Voxel), NSD (%) and ASD (Voxel).

| Networks | Type | Params (M)↓ | FPS↑ | DSC↑ | HD95↓ | NSD↑ | ASD↓ |
|---|---|---|---|---|---|---|---|
| LeViT-UNet++ | C+T | 8.273 | 203.52 | 84.88 | 29.79 | 23.22 | 6.32 |
| Mobile U-ViT | C+T | 2.758 | 216.61 | 80.60 | 50.36 | 19.56 | 14.22 |
| SegFormer3D | T | 4.492 | 221.57 | 84.16 | 37.32 | 21.82 | 8.35 |
| MedNeXt-v1 | C | 5.551 | 148.95 | 85.05 | 47.39 | 23.46 | 12.09 |
| UNet++ | C | 6.980 | 76.59 | 82.62 | 47.72 | 22.11 | 12.45 |

| Networks | Type | Params (M)↓ | FPS↑ | DSC↑ | HD95↓ | NSD↑ | ASD↓ |
|---|---|---|---|---|---|---|---|
| nnUNet | C | 16.539 | 231.99 | 82.46 | 46.48 | 21.52 | 12.23 |
| LeViT-UNet | C+T | 33.867 | 128.47 | 81.50 | 30.20 | 20.58 | 6.15 |
| V-Net | C | 45.598 | 157.18 | 85.82 | 30.68 | 23.91 | 7.18 |
| UXNet | C | 53.006 | 70.54 | 83.87 | 42.90 | 22.80 | 11.09 |
| Swin UNETR | C+T | 62.187 | 61.77 | 83.13 | 44.78 | 22.08 | 11.24 |
| nnFormer | T | 149.464 | 21.81 | 76.22 | 40.99 | 14.96 | 11.16 |

(2) Results on MSD-Heart dataset

We evaluate LeViT-UNet++ on the MSD-Heart dataset for heart segmentation. As shown in **Table 6**, our method obtains a DSC of 88.76, HD95 of 5.20, NSD of 27.67, and ASD of 1.29. Even though nnUNet, UNet++, UXNet, MedNeXt-v1 and Swin UNETR attain higher DSC values, many of them suffer from degraded HD95 and NSD boundary metrics. nnFormer reaches only 4.91 in NSD with a drastically reduced DSC of 52.85, which is not applicable for cardiac segmentation. Relative to the LeViT-UNet baseline, LeViT-UNet++ improves the DSC from 84.26 to 88.76 alongside consistent gains on boundary metrics. Mobile U-ViT performs poorly on both DSC and boundary measurements. This evidence confirms that our LeViT-UNet++ strikes a good balance between volumetric overlap and boundary consistency.

**Table 6.** Quantitative segmentation results of competing methods on the MSD-Heart cardiac segmentation benchmark, including averaged DSC (%), HD95 (Voxel), NSD (%) and ASD (Voxel).

| Networks | DSC ↑ | HD95↓ | NSD↑ | ASD↓ |
|---|---|---|---|---|
| LeViT-UNet++ | 88.76 | 5.20 | 27.67 | 1.29 |
| Mobile U-ViT | 77.03 | 130.53 | 19.41 | 40.15 |
| SegFormer3D | 89.00 | 6.91 | 29.16 | 4.47 |
| MedNeXt-v1 | 90.48 | 14.66 | 32.26 | 1.73 |
| UNet++ | 90.61 | 38.60 | 33.72 | 5.36 |
| nnUNet | 91.03 | 4.48 | 34.93 | 1.56 |
| LeViT-UNet | 84.26 | 46.44 | 24.86 | 7.05 |
| V-Net | 89.03 | 16.73 | 24.45 | 2.64 |
| UXNet | 90.56 | 5.69 | 33.45 | 2.79 |
| Swin UNETR | 90.43 | 5.07 | 32.70 | 1.00 |
| nnFormer | 52.85 | 48.73 | 4.91 | 15.98 |

### 4.5 Ablation study

We perform ablation experiments to quantify the individual contributions of the proposed HWMD and DLGD modules to our LeViT-UNet++. **Table 7** presents the 2D ablation results on the Synapse dataset. Compared with the baseline LeViT-UNet, integrating only the HWMD module substantially improves the DSC, which validates the effectiveness of the Haar-wavelet multi-scale downsampling stem for feature extraction. When equipped with the DLGD decoder alone, the model achieves a remarkable reduction in parameter count while retaining competitive segmentation accuracy.

Combining both modules yields LeViT-UNet++, which achieves the best overall trade-off between segmentation performance, model compactness and inference speed.

**Table 7**. Ablation study to investigate the contributions of the proposed HWMD and DLGD modules on Synpase dataset under 2D setting. We report the number of parameters (Params, in M), inference speed (FPS), averaged DSC (%), and HD95.

| Networks | HWMD | DLGD | Params (M) ↓ | FPS ↑ | DSC ↑ | HD95↓ |
|---|---|---|---|---|---|---|
| LeViT-UNet | × | × | 15.896 | 237.31 | 73.69 | 23.92 |
| +HWMD | √ | × | 15.898 | 221.59 | 77.70 | 25.69 |
| +DLGD | × | √ | 8.022 | 254.03 | 75.33 | 21.51 |
| LeViT-UNet++ | √ | √ | 8.025 | 241.72 | 76.57 | 18.29 |

To further verify the generalizability of these two designs in volumetric segmentation, we carry out the same ablation configuration on the 3D MSD-Heart dataset, as listed in **Table 8**. We obtain consistent conclusions: HWMD improves volumetric segmentation accuracy, while DLGD substantially reduces the number of model parameters. The complete LeViT-UNet++ achieves the best HD95 and ASD metric for boundary segmentation, which demonstrates that HWMD and DLGD are complementary and work synergistically for 3D cardiac segmentation.

**Table 8.** Ablation results for HWMD and DLGD components, with comparisons of model size, inference efficiency and segmentation metrics under 3D setting for MSD-Heart dataset.

| Networks | HWMD | DLGD | Params (M) ↓ | FPS ↑ | DSC↑ | HD95↓ | NSD↑ | ASD↓ |
|---|---|---|---|---|---|---|---|---|
| LeViT-UNet | × | × | 33.859 | 132.84 | 84.26 | 46.44 | 24.86 | 7.05 |
| +HWMD | √ | × | 33.867 | 130.54 | 88.51 | 16.65 | 25.98 | 2.68 |
| +DLGD | × | √ | 8.265 | 192.20 | 88.10 | 5.58 | 28.84 | 1.41 |
| LeViT-UNet++ | √ | √ | 8.273 | 178.67 | 88.76 | 5.20 | 27.67 | 1.29 |

### 4.6 Visualization of segmentation results

In Figures 4 and 5, we qualitatively compare our LeViT-UNet++ against vanilla LeViT-UNet, Mobile U-ViT, SegFormer and Swin UNETR. Figure 4 illustrates 2D slice overlays for Synapse and ACDC, and Figure 5 provides 3D volume renderings for LA2018 and MSD Heart, with the first row of each panel being the ground truth. Our method yields segmentation masks with crisp boundaries and faithful anatomical contours: it avoids the blurred edges and under-segmentation observed in vanilla LeViT-UNet, suppresses the spurious fragments prevalent in Mobile U-ViT, and outperforms SegFormer and Swin UNETR, which often produce distorted outlines and incomplete structures. In 3D reconstruction, LeViT-UNet++ preserves the complete topological morphology of cardiac structures, while the competing models suffer from surface deformation, missing regions and discrete noise. These visual results demonstrate the capacity of our method to capture fine anatomical cues in both 2D and 3D segmentation scenarios.

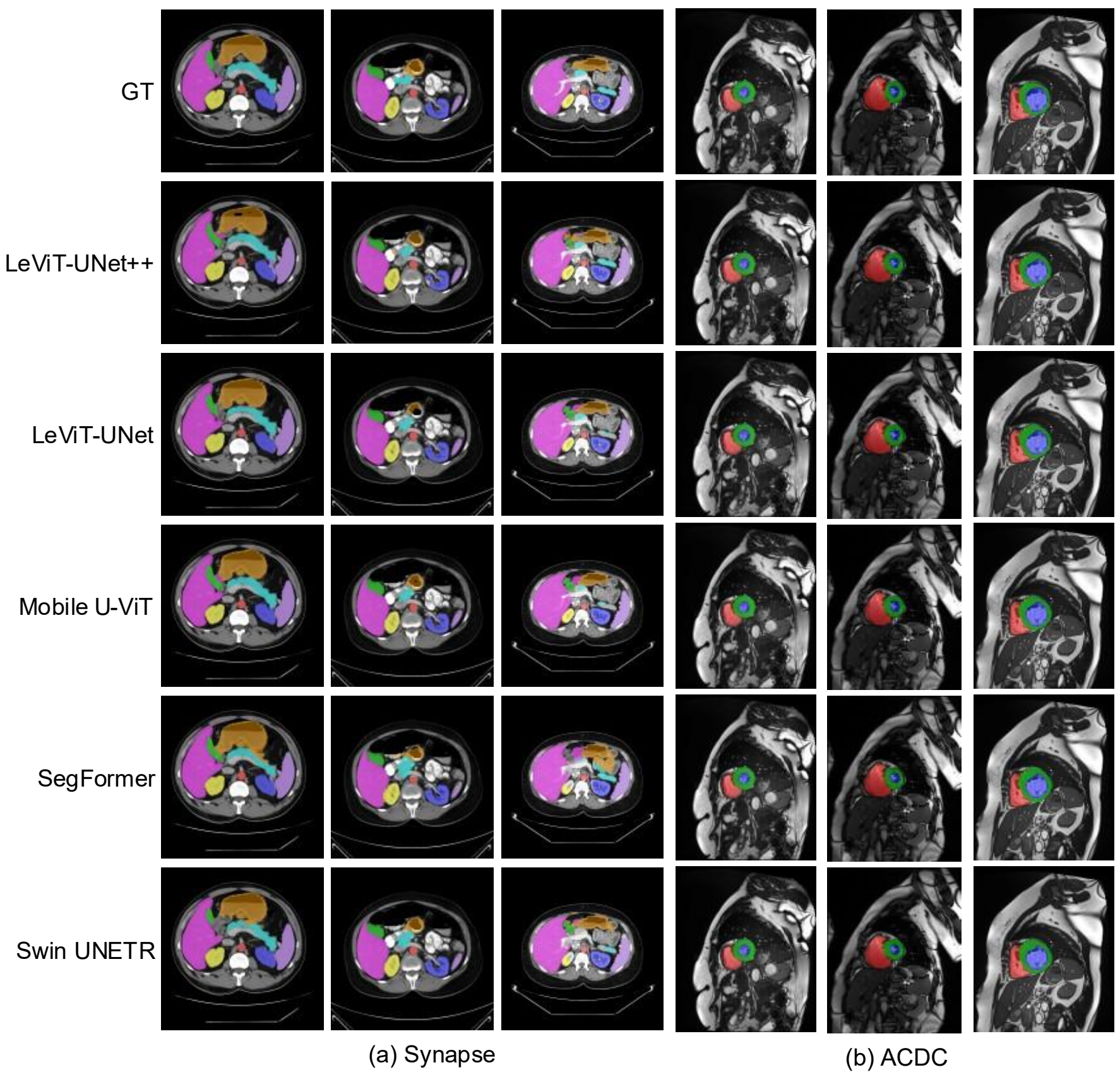

Figure 4. Qualitative visualization of segmentation results for the Synapse and ACDC datasets under the 2D setting.

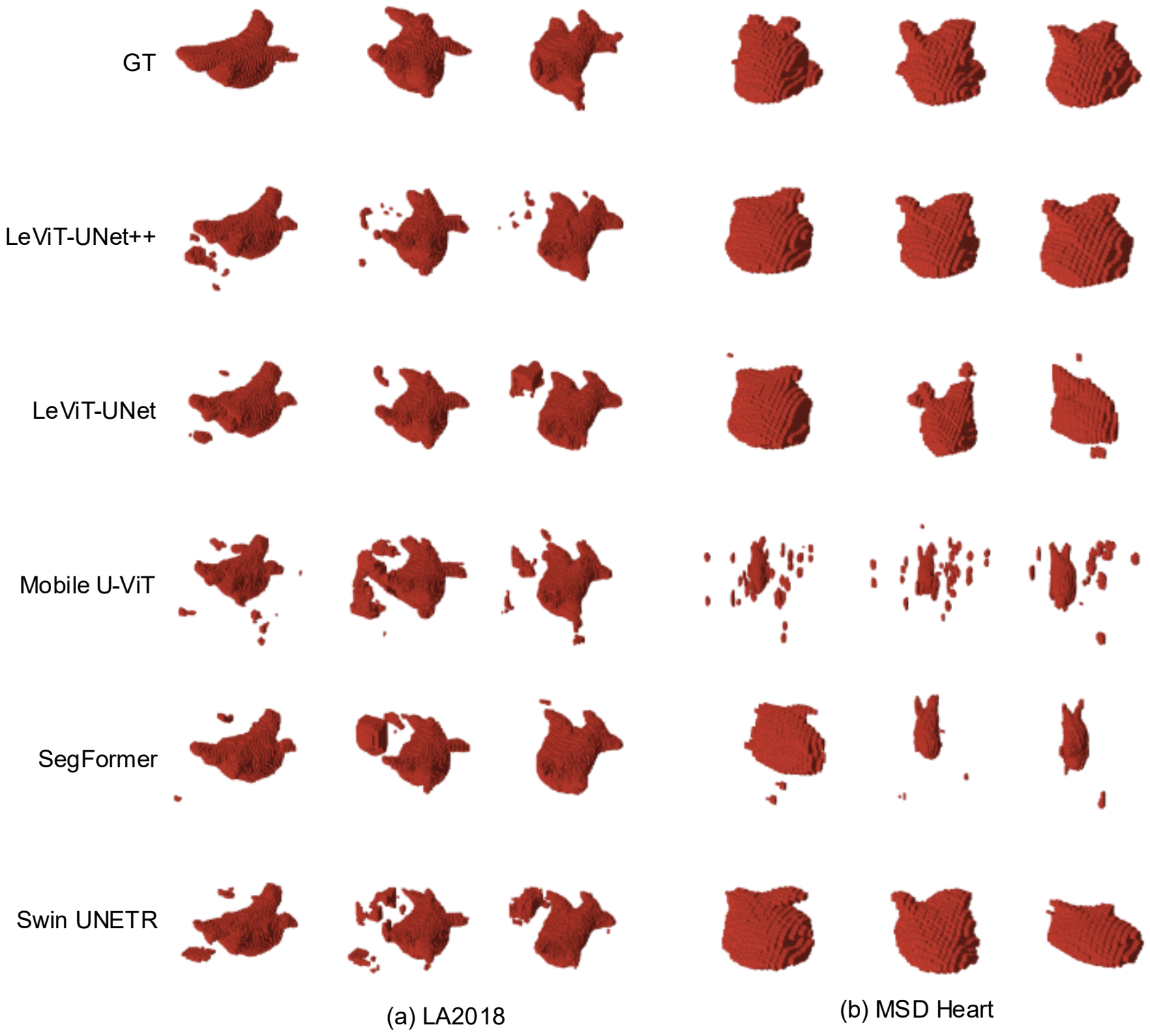

Figure 5. Visualization of segmentation results on the LA2018 and MSD Heart datasets under the 3D setting.

## 5. Discussion

LeViT-UNet++ was developed to improve the balance between segmentation accuracy and computational efficiency by addressing two complementary aspects of feature representation: structural information preservation during downsampling and computationally efficient feature integration during decoding. Specifically, HWMD incorporates Haar wavelet decomposition and frequency-specific multi-scale depthwise convolutions to retain complementary low- and high-frequency representations during the initial downsampling stage. Meanwhile, DLGD employs depthwise separable convolutions to progressively integrate local spatial features and global contextual representations while reducing decoder complexity.

The experimental results across four medical image segmentation benchmarks demonstrate the potential of these complementary designs in both 2D and 3D segmentation settings. Compared with the original LeViT-UNet, the proposed architecture achieves notable improvements on the Synapse and LA2018 datasets while substantially reducing the number of model parameters. In particular, the parameter reduction is primarily attributed to the lightweight decoder design, whereas HWMD contributes to the preservation of complementary structural representations during feature extraction. These results suggest that improving architectural efficiency does not necessarily require sacrificing the representation of fine-grained structural information.

Despite these promising results, the proposed architecture has several limitations that warrant further investigation. First, although LeViT-UNet++ achieves a favorable trade-off between segmentation performance and computational efficiency, its performance does not consistently surpass that of state-of-the-art segmentation methods, particularly those employing more sophisticated feature representation and fusion strategies. This suggests that improving computational efficiency while maintaining or further enhancing segmentation accuracy remains a challenging problem. One possible contributing factor is the limited adaptability of the current multi-scale feature fusion strategy. Specifically, DLGD progressively integrates local and global representations through concatenation-based fusion and skip connections. Although this strategy is computationally efficient, it may not fully capture the complex dependencies and complementary relationships among features at different spatial resolutions. More adaptive fusion mechanisms, such as attention-guided feature selection, content-aware feature aggregation, or learnable cross-scale interactions, may provide more effective ways to exploit complementary local details and global contextual information while preserving the lightweight nature of the architecture. Furthermore, knowledge distillation represents a promising paradigm for transferring knowledge from heavy, large-scale models into lightweight architectures such as our proposed LeViT-UNet++, which helps mitigate the requirement for massive annotated training samples in medical image segmentation. We hope these explorations will inspire further advances toward compact, data-efficient models for routine clinical workflows.

Second, the evaluation was conducted on a limited number of publicly available benchmarks. Although these datasets cover different anatomical structures and imaging modalities, further validation on larger, more diverse, and clinically representative datasets is required to establish the robustness and generalizability of the proposed architecture. In addition, its computational efficiency requires systematic evaluation across diverse hardware platforms and deployment settings, especially under real-world clinical constraints. Furthermore, the current feature fusion strategy mainly relies on concatenation and depthwise convolution, which may restrict its capacity to adaptively model complex interdependencies among multi-scale features. Accordingly, our future work will focus on developing more effective yet computationally economical multi-scale feature fusion schemes, extending evaluations to larger clinical datasets, and exploring the feasibility of deploying LeViT-UNet++ in resource-limited medical imaging scenarios.

## 6. Conclusion

This study presents LeViT-UNet++, a lightweight convolution–Transformer hybrid framework for medical image segmentation, aiming to balance segmentation accuracy and computational efficiency. It integrates the Haar Wavelet Multi-Scale Downsampling (HWMD) module to preserve fine-grained structural information and the Depthwise-based Local and Global feature fusion and learning Decoder (DLGD) to efficiently combine local spatial details with global contextual representations. Extensive experiments across four benchmarks (Synapse, ACDC, LA2018, and MSD-Heart) under both 2D and 3D settings demonstrate that LeViT-UNet++ delivers competitive segmentation performance with substantially reduced model complexity and favorable inference speed. These findings highlight the effectiveness of combining wavelet-based structural preservation with efficient local–global feature fusion, offering a promising solution for medical image segmentation in resource- and latency-constrained environments.

## Acknowledge

This work is supported by the the Fundamental Research Funds for the Central Universities (JLU), Guangdong Provincial Key Laboratory of Human Digital Twin (No. 2022B1212010004), and the Open Project Program of the State Key Laboratory of CAD&CG Zhejiang University (No. A2322). We would like to thank Xuan Zhang, Xinglong Wu, and Yajun Yu for their insightful and helpful discussions.

## Conflict of interest statement

The authors declare no conflicts of interest.